\documentclass{svproc}
\usepackage{url}

\usepackage{graphicx} 
\graphicspath{{Figures/}} 
\usepackage{amsmath,amssymb}

\begin{document}
\mainmatter              
\title{C-space Analysis using Tropical Geometry}
\titlerunning{C-space Analysis using Tropical Geometry}  
%
\author{Abhilash Nayak}
\authorrunning{Abhilash Nayak} 
%
%
\institute{Laboratoire des Sciences du Numérique de Nantes~(LS2N), France.\\
\email{abhilash.nayak@ls2n.fr}
}

\maketitle              

\begin{abstract}
Configuration space~(C-space) of a mechanism is a real variety describing the set of feasible configurations that it can attain. To understand the behavior of a mechanism, it is crucial to identify and scrutinize especially the singular points of its C-space. They usually appear when the variety intersects itself, leading to different branches of motion. There exist many approaches to detect those intersections if they are transversal. However, the problem remains challenging if there are tangential, cuspidal, inter-dimensional or a combination of these intersections. This paper exploits an approach acquired from tropical geometry to analyze the neighborhood of any point on C-spaces of 1-degree-of-freedom~(\emph{dof}) mechanisms. This is done by finding the approximate rational parametrization of the curve(s) passing through the given point using Puiseux series. The proposed approach is shown to succesfully detect the transversal branchings in two foldable four bar mechanisms and a cusp in the configuration curve of the double Watt mechanism.
\keywords{kinematics, configuration space, singularity analysis, tropical geometry, Puiseux series}
\end{abstract}
\section{Motivation}
Often, a mechanism can possess different branches of motion~(also known as motion phases, motion modes or operation modes), which correspond to sub-varieties of its C-space. A mechanism is in a singularity where the branching occurs in the configuration space. The behavior of a mechanism at these singularities is usually studied by examining the local tangent space or tangent cone of the analytic/algebraic variety describing its C-space~\cite{Muller2019singular}. Although research in this regard is aplenty, the target has been on mechanisms that exhibit \emph{transversal} intersections of the manifolds at a singularity, where the tangent spaces corresponding to each branch are distinct and well defined. A reason for this might be the scarcity of mechanisms with non-transversal intersections between their branches of motion.\\
However, recently, L{\'o}pez-Custodio et al. proposed a novel approach to design 1-\emph{dof} mechanisms with cuspidal~\cite{Custodio2019} or tangential~\cite{Custodio2020} intersection of branches of motion. Furthermore, they presented a method to detect and analyze tangential branches of motion. The problem arises when there is a cusp in the C-space as tangents are not defined there.\\
This paper derives some ideas from tropical geometry and puts forth a methodology to find the local approximation of the C-space of 1-\emph{dof} mechanisms. It aims to detect and analyze the curves passing through any point on the C-space including and especially cusps. A brief outline of the proposed C-space analysis is as follows. Study's kinematic mapping~\cite{Husty2007} is used to derive the constraint equations of the given mechanism. Thus, the corresponding algebraic variety describes the C-space in terms of Study parameters unlike the joint parameters used in L{\'o}pez-Custodio et al.~\cite{Custodio2019,Custodio2020}. Then, the singularities are determined by examining the tangent space. Consequently, the Pusieux series approximation~(introduced in Section~\ref{sec:Puiseux})  of the of curve(s) branching from a singularity is determined using the commutative algebra system \textsc{Singular}~\cite{Greuel2009} and the program \verb|Gfan|~\cite{gfan} to know the type of branching. Finally, the proposed approach is applied on two four bar linkages with transversal intersections of branches of motion in Section~\ref{sec:4bar} and on a double Watt mechanism with a cusp in its C-space in Section~\ref{sec:DWatt}.

\section{Puiseux power series} \label{sec:Puiseux}
The field $K\{\{x\}\}:=\cup_{n=1}^{\infty}K((x^{1/n}))$ of \emph{power series} such that a polynomial is of the form $\sum_{i=0}^{\infty}a_ix^i$, where $a_i \in K$ is called the \emph{Puiseux series}. It is the series with fractional exponents.\\
The Newton-Puiseux algorithm was first proposed by Newton in the 70's which was quite forgotten but resurrected by Puiseux in the early 19th century. Given a polynomial $f(x,y)$, Newton-Puiseux algorithm can be used to compute a power series expansion for $y$ by viewing it as solving a polynomial equation in $y$ with coefficients in $K\{\{x\}\}$. It does so by looking at each term and searching for conditions for cancellation of coefficients of lowest order. This is illustrated through an example. Let 
\begin{equation*} 
	f(x,y)=4y^3+4xy^2+x^2y+2x^4.
\end{equation*}
If a Puiseux series of the form $y(x)$ has to be constructed about the origin~$(x=y=0)$, it should have the following form:
\begin{align*} 
\bar{y}(x)=\, & c_1x^{\gamma_1}+c_2x^{\gamma_1+\gamma_2}+ c_3x^{\gamma_1+\gamma_2+\gamma_3}+ \dots \nonumber = x^{\gamma_1}(c_1+y_1(x))
\end{align*}
Substituting $\bar{y}(x)$ in $f(x,y)$ yields
\begin{align*}
f(x,\bar{y}(x))	= 2x^4+x^{2+\gamma_1}c_1+4x^{1+2 \gamma_1}c_1^2+4x^{3 \gamma_1}c_1^3+g(x,y_1)
\end{align*}
Since, the series is to approximate $f(x,y)$ around the origin, $f(x,\bar{y})$ must have a factor $x$ or its powers. Hence, the necessary condition for $f(x,\bar{y})=0$ is that the terms of lowest order must vanish. Therefore, at least two terms must have the same order and it must not be greater than the order of any other. Besides, all terms of the smallest order must vanish.\\
Thus, the solution to $\gamma_1$ such that the set $\min \{4, 2+\gamma_1, 1+2\gamma_1, 3\gamma_1\}$ contains at least two terms is $\gamma_1=\{2,1\}$.
For $\gamma_1=2$, $c_1+2=0 \implies c_1=-2$. Therefore, the terms of $f$ corresponding to the lowest orders are $2x^4+x^2y$ which gives the first term of the Puiseux series, $y=-2x^2$. For $\gamma_1=1$, the terms of the lowest orders are $2x^4+x^2y+4xy^2$ yielding the first order Puiseux series, $y=-(\frac{1}{2})x$. Iterating this procedure may lead to higher order approximations.\\  
The extension of Newton-Puiseux algorithm to higher dimensions lies under the framework of \emph{tropical geometry}, which is a piecewise-linear version of \emph{algebraic geometry}~\cite{Maclagan2015}. Its geometry is the \emph{tropical semiring} ($\mathbb{R} \cup \infty $), where the operations are: \emph{tropical sum}, which is the minimum of two elements and \emph{tropical product}, which is the usual sum.\\
Let $f=\sum c_\mathbf{a} \mathbf{x}^\mathbf{a}$ be a polynomial in $\mathbb{C}^n$ with $\mathbf{x}^\mathbf{a}=x_1^{a_1}x_2^{a_2}...x_n^{a_n}$.
Then, the \emph{initial form} of $f$ with respect to a vector $\boldsymbol{\gamma} \in \mathbb{R}^n$ is defined by
\begin{equation}\label{eq:in}
	\text{in}_{\boldsymbol{\gamma}}(f)=\sum_{\langle\mathbf{a},\boldsymbol{\gamma}\rangle=M} c_\mathbf{a} \mathbf{x}^\mathbf{a}.
\end{equation}
where $\langle\cdot,\cdot\rangle$ is the scalar product and $M=\min\{\langle\mathbf{a},\boldsymbol{\gamma}\rangle: c_\mathbf{a} \neq 0\}$. 
The \emph{tropical variety} of $f$, trop($V(f)$) is the set of $\boldsymbol{\gamma}$ for which $\text{in}_{\boldsymbol{\gamma}}(f)$ consists of at least two monomials. In the previous example, trop$(V(f))=\{(1,2),(1,1)\}$ with $\text{in}_{(1,2)}=2x^4+x^2y$ and $\text{in}_{(1,1)}=2x^4+x^2y+4xy^2$ which are useful to find the Puiseux series.\\
Hence, tropical variety is the image of an algebraic variety over the Puiseux series and the tropical lifting algorithm~\cite{gfan} gives the \emph{Puiseux valued lift} of a point in the algebraic variety.

\section{Foldable four bar mechanisms} \label{sec:4bar}
\begin{figure}[t]
	\centering
	\begin{minipage}{0.35\textwidth}
	\centering
\includegraphics[width=\textwidth]{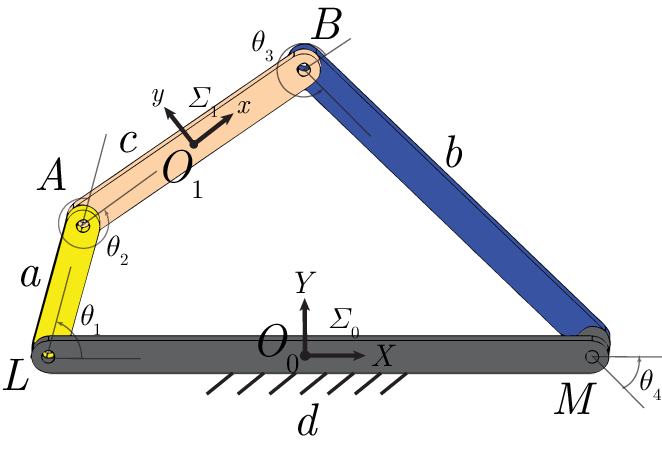}
\caption{A four bar mechanism}\label{fig:4bargen}
	\end{minipage}%
	\begin{minipage}{0.65\textwidth}
	\centering
\includegraphics[width=\textwidth]{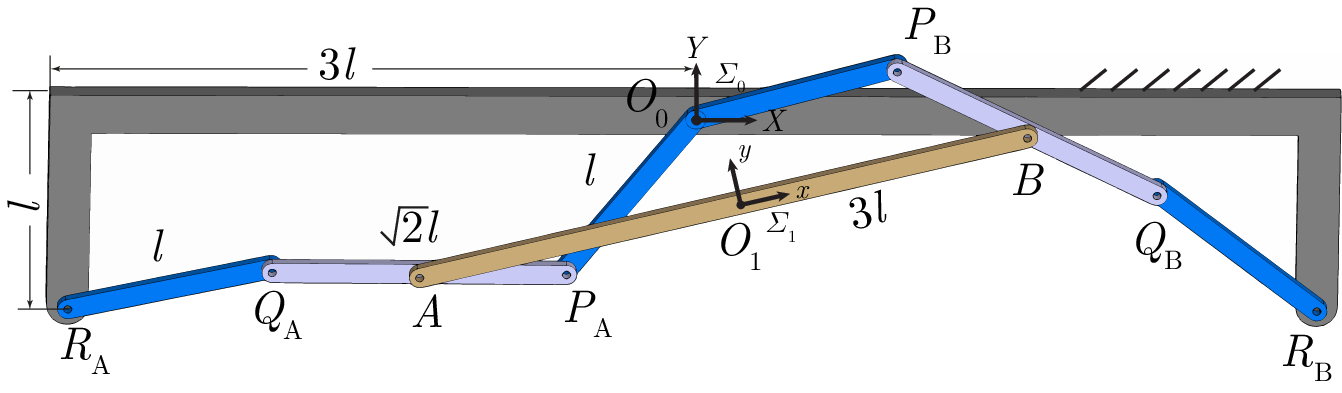}
\caption{Double Watt mechanism}\label{fig:dWg}
	\end{minipage}%
\end{figure}

A four bar mechanism shown in Fig.~\ref{fig:4bargen} is a single loop planar mechanism consisting of a fixed link, three moving links~(of lenghts $d,a,b,c$, respectively) and four revolute joints with parallel axes exhibiting 1-\emph{dof}. 
 To derive the algebraic constraint equations, a fixed coordinate frame $\Sigma_0$ is attached to the fixed link $LM$ and a moving coordinate frame $\Sigma_1$ is attached to the coupler $AB$. The mechanism can be split into two serial 2R~(R represents a revolute joint) linkages $O_0LAO_1$ and $O_0MBO_1$ and the transformation matrix between $\Sigma_0$ and $\Sigma_1$ can be written as follows~\cite{Husty2007} :
 \begin{subequations}
  \begin{align}
 \mathbf{T}_L&=\mathbf{G}_1 \mathbf{M}_1 \mathbf{G}_2 \mathbf{M}_2 \mathbf{G}_3 \label{eq:fw} \\
 \mathbf{T}_R&=\mathbf{G}^{-1}_1 \mathbf{M}^{-1}_4 \mathbf{G}^{-1}_4 \mathbf{M}^{-1}_3 \mathbf{G}^{-1}_3 \label{eq:bw}
 \end{align}
 \end{subequations}
with $\mathbf{T}_L\mathbf{T}_R=\mathbf{I}$ (this relation can be used to derive the constraint equations in the joint space). Matrices $\mathbf{G}_{i}$ are fixed transformations consisting of link lengths while $\mathbf{M}_i$ are functions of joint angles $\theta_i$. After Weierstrauss substitution $v_i=\tan\left(\frac{\theta_i}{2}\right)$, $\mathbf{T}_L$ can be converted to Study parameters\footnote{Study's kinematic mapping maps elements of SE(3) to points $(x_i,y_i) \in \mathbb{P}^7,\,i=0,1,2,3$. $x_i,y_i$ are called Study parameters. Similarly, the mapping from SE(2) to points $(x_0,x_3,y_1,y_2) \in \mathbb{P}^3$ is known as Blaschke-Gr\"unwald mapping. It can be derived from Study's kinematic mapping by setting $x_1=x_2=y_0=y_3=0$.}
\begin{align}
	x_0&=4v_{{1}}v_{{2}}-4,\quad  y_1=( 2a-c+d ) v_{{1}}v_{{2}}+2a+c-d,  \\
	x_3&=-4v_{{1}}-4v_{{2}}, \quad 
	 y_2=( 2a+c+d) v_{{1}}+ ( -2a+c+d) v_{{2}}.
\end{align}
Using Linear Implicitization Algorithm~(LIA)~\cite[Chapter~4]{Muller2019singular} for planar kinematics, $v_1$ and $v_2$ can be eliminated to obtain a constraint equation~\eqref{eq:L}. Similarly, another constraint equation~\eqref{eq:R} can be obtained corresponding to $\mathbf{T}_R$.
\begin{subequations}
	\begin{align}
 f_L:= & ( c-d+2\,a )  ( -c+d+2\,a ) {x_{{0}}}^{2}+
( -8\,c+8\,d ) x_{{0}}y_{{1}}  + ( c+d+2\,a )\nonumber \\  & 
( -c-d+2\,a ) {x_{{3}}}^{2}   + ( -8\,c-8\,d ) x_{
	{3}}y_{{2}}-16\,{y_{{1}}}^{2}-16\,{y_{{2}}}^{2} =0, \label{eq:L} \\
f_R:= &  ( c-d+2\,b )  ( -c+d+2\,b ) {x_{{0}}}^{2}+
( 8\,c-8\,d ) x_{{0}}y_{{1}}  + ( c+d+2\,b ) \nonumber \\ &
( -c-d+2\,b ) {x_{{3}}}^{2}+ ( 8\,c+8\,d ) x_{{
		3}}y_{{2}} -16\,{y_{{1}}}^{2}-16\,{y_{{2}}}^{2} =0. \label{eq:R}
	\end{align} 
\end{subequations} 
When Eqs.~\eqref{eq:L} and~\eqref{eq:R} are normalized with $x_0=1$ or $x_3=1$, it is well known that they represent two hyperboloids in the kinematic image space~$\mathbb{P}^3$. Their intersection is the C-space of the mechanism which is the variety $f_L=f_R=0$.\\
Essentially, the mechanism is foldable when its configuration curve in $\mathbb{C}^3$ has a self intersection point, where the tangent planes of the hyperboloids coincide. This can only happen when the link lengths satisfy certain relations. To locally analyze the C-space in a singularity, let us consider two foldable four bar mechanisms given by $a+c=b+d$ and $c=a+b+d$. In both cases, the normalization condition $x_0=1$ is used so that the rest of the analysis is in $\mathbb{C}^3$.
\subsection{Case 1. $a+c=b+d$}
Equations~\eqref{eq:L} and~\eqref{eq:R} are simplified by fixing the link lengths $a=1,b=2,c=4,d=3$. This gives the constraint ideal~\cite{Cox2007}:
\begin{align*}
\mathcal{I}_1=\langle f_L, f_R \rangle = \langle & 3-8\,y_{{1}}-45\,{x_{{3}}}^{2}-56\,x_{{3}}y_{{2}
}-16\,{y_{{1}}}^{2}-16\,{y_{{2}}}^{2},
 \\ & 15+8y_{{1}}-33\,{x_{{3}}}^{2}+56\,x_{{3}}y_{{2
 }}-16\,{y_{{1}}}^{2}-16\,{y_{{2}}}^{2}
\rangle.
\end{align*}
It is known that when this four bar mechanism is folded, the rotation matrix between $\Sigma_0$ and $\Sigma_1$ is the identity, implying $x_0=1, x_3=0$. Substituting it in $\mathcal{I}_1$ leads to singular point $S_1=(0,-\frac{3}{4},0)$. In fact, the forward Jacobian matrix can also be calculated ~\cite[Chapter~4]{Muller2019singular} whose kernel yields two tangents at this point due to the transversal intersection. Nonetheless, the Puiseux series approximation of the curves passing through $S_1$ is determined by first calculating the tropical variety of $\mathcal{I}_1$ at $S_1$, $\text{trop}(V(\mathcal{I}_1))=\{(1,2,1)\}$. It is lifted in \textsc{Singular}~\cite{Greuel2009} to obtain the following first order approximations:
\begin{equation*}
x_3=t,\;y_1=\frac{-28m-3}{4}t^2, \; y_2=mt \;\textrm{with } m= -{\frac {21}{4}} \pm 2\,\sqrt {6} .
\end{equation*}
Higher order approximations can be determined if necessary but it already proves the existence of two curves branching from the singular point. Figure~\ref{fig:fb1} shows the configuration curve as an intersection of two hyperboloids along with its second order approximation at the bifurcation. Also the top view of the mechanism at this singularity is shown. 
\subsection{Case 2. $c=a+b+d$}
\begin{figure}[t]
	\centering
	\begin{minipage}{0.5\textwidth}
		\centering
		\includegraphics[width=0.6\textwidth]{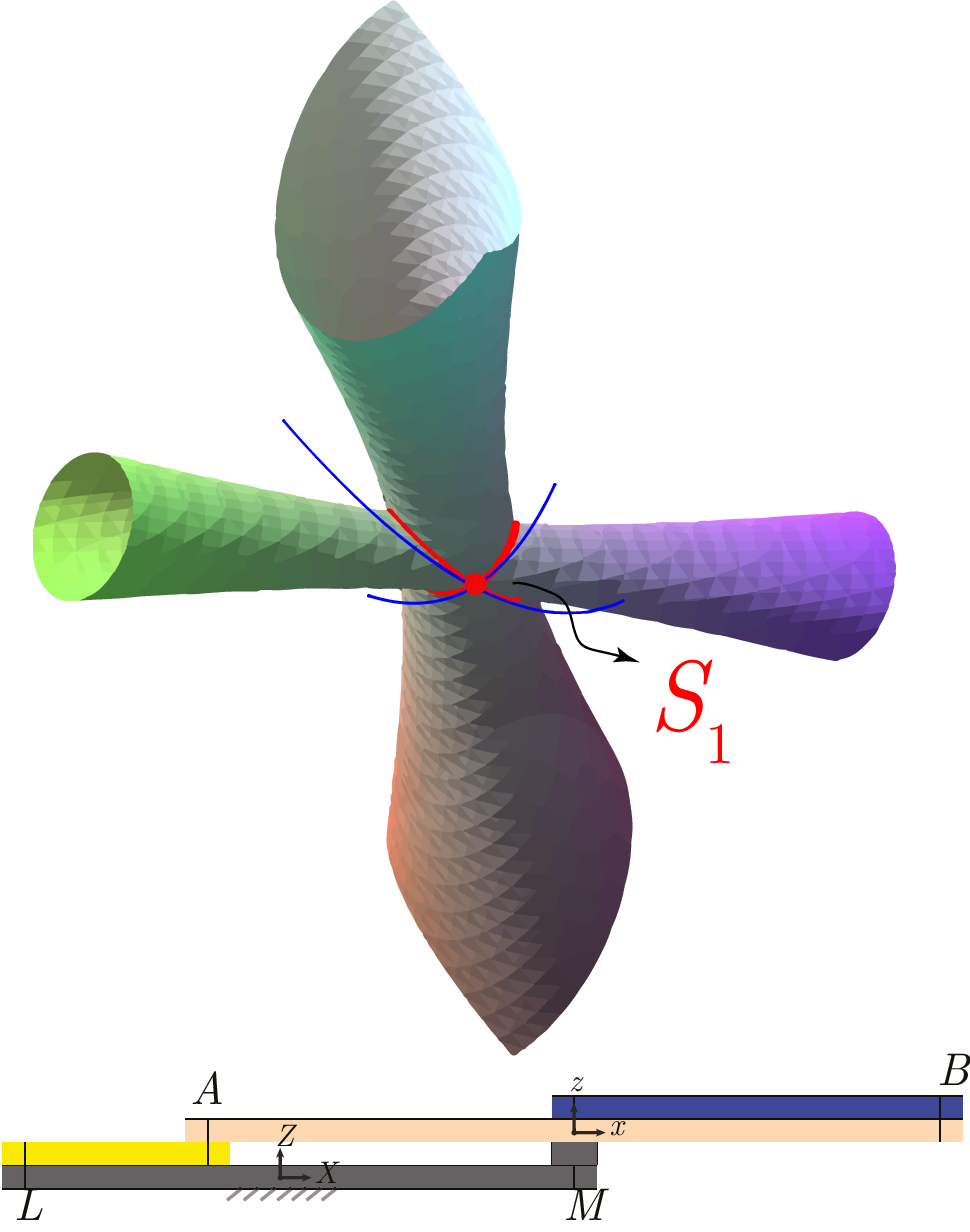}
		\caption[]{a+c=b+d}\label{fig:fb1}
	\end{minipage}%
	\begin{minipage}{0.5\textwidth}
		\centering
		\includegraphics[width=0.7\textwidth]{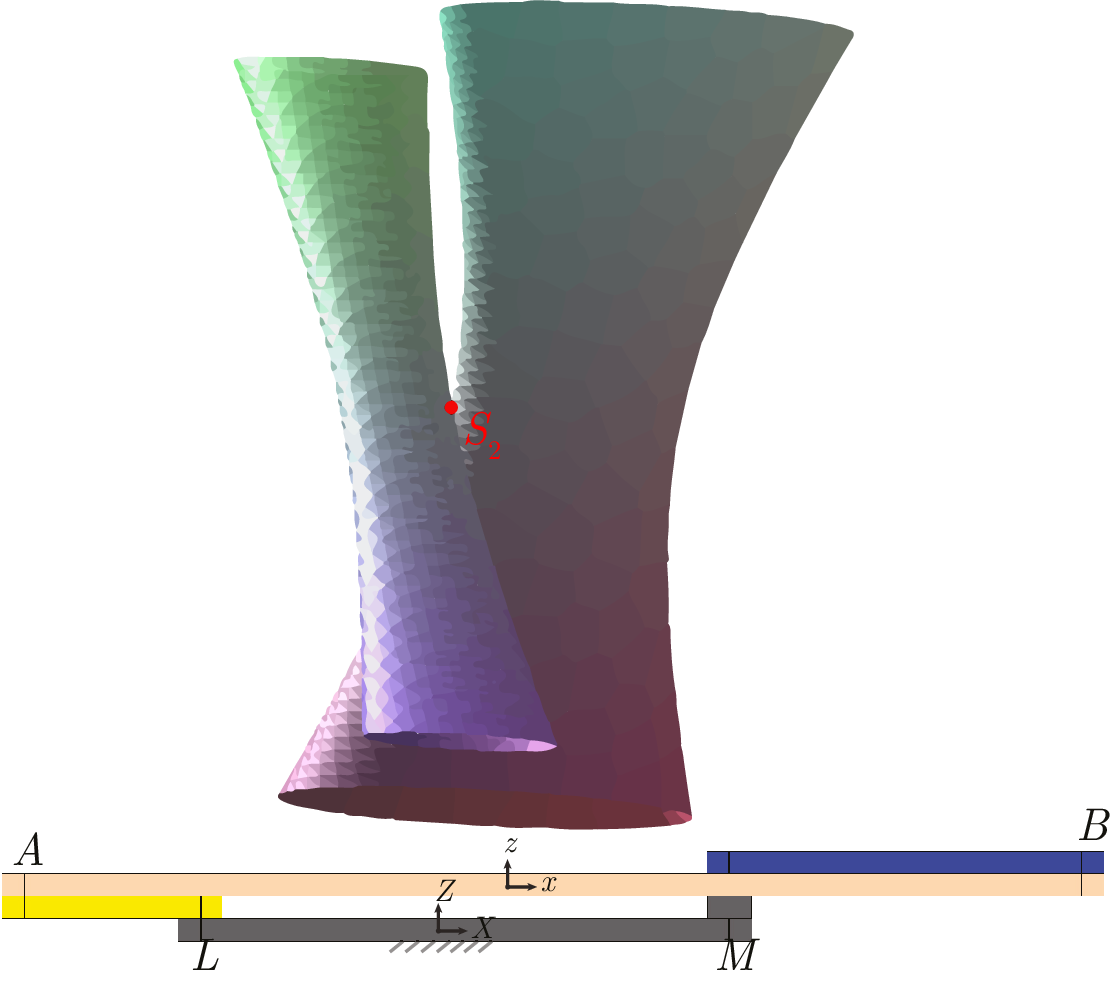}
		\caption[]{c=a+b+d}\label{fig:fb2}
	\end{minipage}%
\end{figure}

Equations~\eqref{eq:L} and~\eqref{eq:R} are simplified by substituting $a=1,b=2,c=6,d=3$. This gives the constraint ideal:
\begin{align*}
\mathcal{I}_2=\langle f_L, f_R \rangle = \langle &-5-24\,y_{{1}}-77\,{x_{{3}}}^{2}-72\,x_{{3}}y_{{
		2}}-16\,{y_{{1}}}^{2}-16\,{y_{{2}}}^{2},
\\ & 7+24y_{{1}}-65\,{x_{{3}}}^{2}+72\,x_{{3}}y_{{2
}}-16\,{y_{{1}}}^{2}-16\,{y_{{2}}}^{2}
\rangle.
\end{align*}
In this case, the singular point turns out to be $S_2=(0,-\frac{1}{4},0)$ and the first order approximation of the branches there is as follows: 
\begin{equation*}
x_3=t, y_1=\frac{-12m-1}{4}t^2, \, y_2=mt \;\textrm{with } m= -{\frac {3}{4}} \pm 2\,i.
\end{equation*}
It shows that there are no real branchings and hence the mechanism cannot physically move out of its only configuration although it could be shaky~\cite[Chapter~5]{Muller2019singular}. Its C-space and the corresponding configuration~(top view) is shown in Fig.~\ref{fig:fb2}.

\section{Double Watt mechanism} \label{sec:DWatt}
\begin{figure}[t] 
	\centering
	\includegraphics[width=\textwidth]{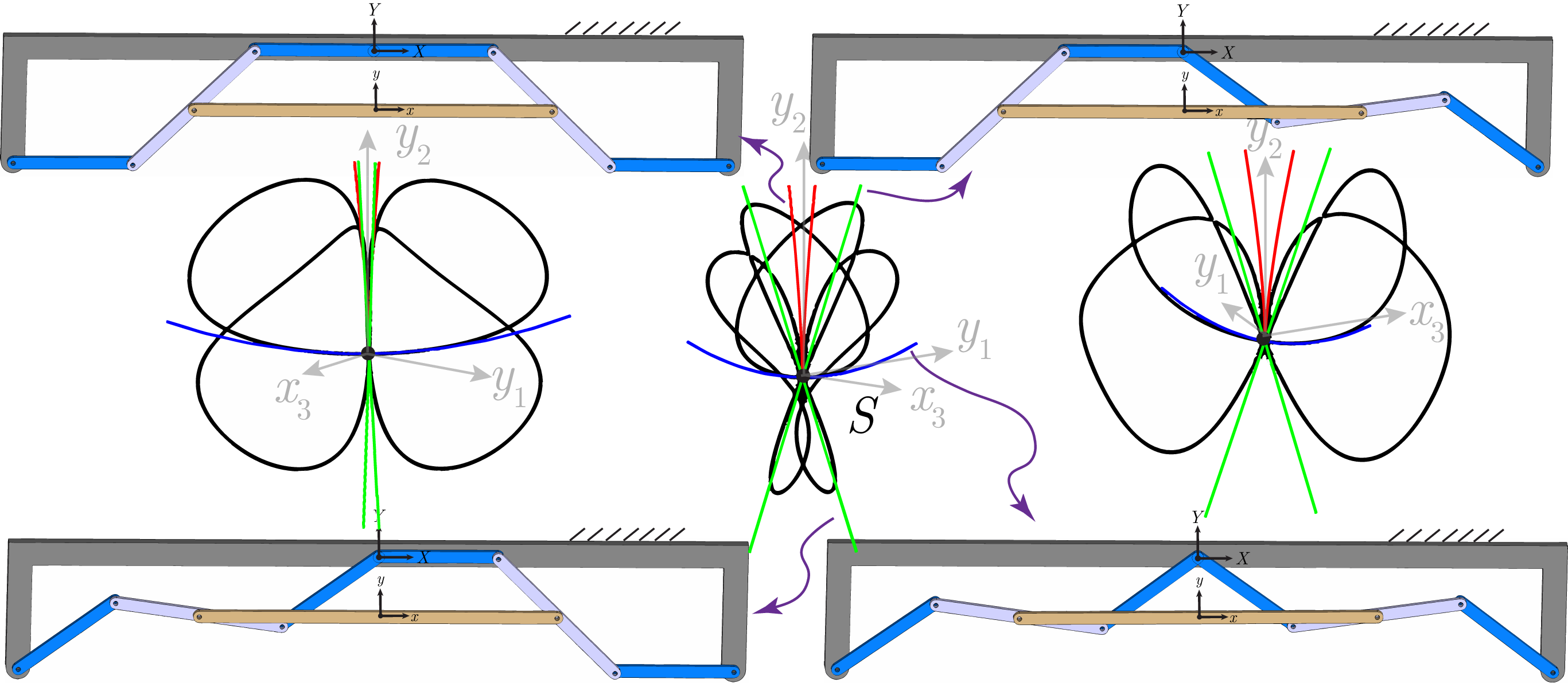}
	\caption[]{C-space of the double Watt mechanism}\label{fig:cd}
\end{figure}

As shown in Fig.~\ref{fig:dWg}, the Connelly Servatius' double Watt mechanism is composed of two four bar Watt linkages, the centers of whose couplers are connected by another link $AB$. It is the first mechanism known to possess a cusp in its C-space~\cite{Custodio2019,Custodio2020}. Knowing that the tangent cone analysis of its C-space fails at the cusp, it will be shown how the Puiseux series can detect the cusp.\\
Similar to Eqs.~\eqref{eq:L} and~\eqref{eq:R}, let the transformation matrix between $\Sigma_0$ and $\Sigma_1$ traversed through four 3R serial linkages $O_0P_AAO_1$, $O_0R_AAO_1$, $O_0P_BBO_1$ and $O_0R_BBO_1$ be $\mathbf{T}_{A_L}, \mathbf{T}_{A_R}, \mathbf{T}_{B_L}$ and $\mathbf{T}_{B_R}$, respectively. As a consequence, the three loop-closure equations in terms of joint angles are $\mathbf{T}_{A_L}(\mathbf{T}_{A_R})^{-1}=\mathbf{T}_{B_R}(\mathbf{T}_{B_L})^{-1}=\mathbf{T}_{A_L}(\mathbf{T}_{A_R})^{-1}=\mathbf{I}$. It results in 9 equtions in 10 variables since the mechanism consists of 10 joints. The inverse Jacobian matrix~\cite[Chapter~4]{Muller2019singular} can be calculated otaining the following singular point:
\begin{align*}
	v_{R_A} = v_{O_0} = v_{R_B} = 0,\, v_{P_A} = v_{P_B} = -\sqrt{2}+1, v_{Q_A} =  v_{Q_B} =  v_A = v_B = \sqrt{2}-1.
\end{align*}
Substituting this point in the Jacobian and determining the kernel does not give any information about the cusp. Moreover, the system of equations is not so simple to deduce the global behaviour of the mechanism. 
Therefore, constraint equations are derived in terms of Study parameters. LIA can be used for this purpose. Instead, a shortcut is to use Eqs.~\eqref{eq:L} and~\eqref{eq:R} that already describe the motion of the coupler of a four bar linkage. Attaching a revolute joint to the coupler corresponds to replacing the Study parameters in the equations by those that are rotated by a given joint angle in $\mathbb{P}^3$~\cite{Husty2007}. Thus, the constraint equations of the two branches look like
\begin{equation} \label{eq:dwcon}
	\begin{split}
	\left.
	\begin{array}{l}
	g_{A_L}=c_2v_A^2+c_1v_A+c_0\\
	g_{A_R}=c'_2v_A^2+c'_1v_A+c'_0 
	\end{array}
	\right\}& \xrightarrow[\text{}]{\text{eliminate $v_A$}} g_A,\; \deg(g_A)=6,\\
	\left.
	\begin{array}{l}
	g_{B_L}=d_2v_B^2+d_1v_B+d_0\\ 
	g_{B_R}=d'_2v_B^2+d'_1v_B+d'_0 
	\end{array}
	\right\}& \xrightarrow[\text{}]{\text{eliminate $v_B$}} g_B,\; \deg(g_B)=6,
	\end{split}
\end{equation}
 where $c_i,c'_i, d_i,d'_i, i=0,1,2$ are functions of $x_0,x_3,y_1,y_2$ and $v_A=\tan\left(\frac{\theta_A}{2}\right)$, $v_B=\tan \left(\frac{\theta_B}{2}\right)$, $\theta_A$ and $\theta_B$ being the joint angles at $A$ and $B$ respectively.\\
Hilbert dimension~\cite[Chapter~9]{Cox2007} of $V(\mathcal{J})$ with $\mathcal{J}=\langle  g_A, g_B\rangle$ is 1 proving that it is a 1-\emph{dof} mechanism (at least in the complex domain). The configuration curve is shown in black in Fig.~\ref{fig:cd} from three different angles about the vertical axis. The primary decomposition~\cite[Chapter~4]{Cox2007} of $\mathcal{J}$ is too difficult to compute and hence does not reveal any property of the C-space. Knowing the singular point in the joint space, the corresponding normalized Study parameters are calculated as $\{x_0=1,x_3=0,y_1=0,y_2=\frac{1}{4} \}$. Since, $x_0 \neq 0$ at the singularity, $\mathcal{J}$ is further simplified by substituting $x_0=1$ and $l=1$ resulting in the singular point $S=(0,0,\frac{1}{4})$. The kernel of the forward Jacobian matrix at $S$ spans the whole space $\mathbb{C}^3$ whereas at a regular point, it is a tangent line. To this end, the only information that can be incurred is that the C-space should have multiple curves passing through the singular point than just the cusp.\\
The tropical variety of $\mathcal{J}$ at $S$ calculated in \verb|Gfan| yields the following six vectors $\boldsymbol{\gamma}_i, i=1,\ldots,6.$
\begin{equation*}
	\text{trop}(V(\mathcal{J}))=\{(2,2,1),(1,1,2),(1,1,0),(1,0,1),(1,0,0),(1,1,1)\}.
\end{equation*}
\textsc{Singular} and Maple are used to lift $\boldsymbol{\gamma}_i$ to obtain the Puiseux series approximations of the curves passing through $S$. The non-trivial tropical liftings are as follows:
\begin{itemize}
	\item $\boldsymbol{\gamma}_2=(1,1,2)$: The initial forms according to Eq.~\eqref{eq:in} are\\
	$\textrm{in}_{\boldsymbol{\gamma}_2}( g_A)=76\,{x_{{3}}}^{3}+288\,{x_{{3}}}^{2}y_{{1}}-64\,x_{{3}}{y_{{1}}}^{2}+
	16\,x_{{3}}y_{{2}}-64\,y_{{1}}y_{{2}}$ and\\ $\textrm{in}_{\boldsymbol{\gamma}_2}( g_B)=-64\, \left( 7\,x_{{3}}+8\,y_{{1}} \right)  \left( x_{{3}}-4\,y_{{1}}
	\right) $.\\
	Solving them gives $\{y_1=-\frac{7}{8}x_3,\,y_2=\frac{25}{8}x_3^2\}$. Thus, the space curve with a parametric representation $[t, -\frac{7}{8}t,\,y_2=\frac{25}{8}t^2]$ is the first order approximation of one of the curves passing through $S$, shown in blue in Fig.~\ref{fig:cd}. It is noteworthy that the degrees of the rational variable $x_3$ are $\{1,1,2\}$, which are in compliance with the  tropical variety. The corresponding configuration of the mechanism is also depicted and the joint angles at $A$ and $B$ are given by $v_A=v_B=7+5\sqrt{2}$. It is not a singular configuration in the joint space and the mechanism can move in and out smoothly through $S$.
	\item $\boldsymbol{\gamma}_6=(1,1,1)$: The initial forms according to Eq.~\eqref{eq:in} are\\
	$\textrm{in}_{\boldsymbol{\gamma}_6}( g_A)=16y_2(x_3-4y_1)$ and\\ $\textrm{in}_{\boldsymbol{\gamma}_6}( g_B)=-64\, \left( 7\,x_{{3}}+8\,y_{{1}} \right)  \left( x_{{3}}-4\,y_{{1}}
	\right) $.\\
	Solving them gives $\{y_1=-\frac{7x_3}{8},\,y_2=0\}$ (a special case of the aforementioned lifting) and
	$\{y_1=-\frac{x_3}{4},\,y_2=y_2\}$. As $y_2$ can take any value, its Puiseux series in terms of $x_3$ at $S$ can be determined by substituting $y_1=-\frac{x_3}{4}$ in $ g_A$ and $ g_B$ of Eq.~\eqref{eq:dwcon}. The resulting two equations are bivariate and hence the \verb|algcurves[puiseux]| command of Maple can be used to determine the planar Puiseux series. It leads to four first order terms of which two are complex, one is a constant~(can be discarded as $\deg(x_3)=0 \neq 1$ and it does not comply with $\boldsymbol{\gamma}_6$) and the other one is $\frac{(6x_3)^{\frac{2}{3}}}{2\sqrt [3]{3}}$. The latter is a cusp and its parametric representation can be simplified as $[t^3, -\frac{t^3}{64},\frac{3t^2}{2}]$ which yields both branches of the cusp passing through $S$ shown as the red curves in Fig.~\ref{fig:cd}. This is a singularity of the C-space, irrespective of its parametrization in terms of Study parameters or joint angles.\\
	Finding a second order approximation yields another series $y_1=35x_3^2/8$. To comply its degree with $\boldsymbol\gamma_6$, its square root must be considered and the final branching can be parametrized as $[t, -\frac{t}{4},\pm \sqrt{\frac{35}{8}}t]$ shown as the green curves in Fig.~\ref{fig:cd}. This is not a singularity either.\\
	The rest of the elements of $\text{trop}(V(\mathcal{J}))$ i.e. $\boldsymbol\gamma_1,\boldsymbol\gamma_3,\boldsymbol\gamma_4,\boldsymbol\gamma_5$ yield either complex series or are special cases of the ones already described.
\end{itemize}
\section{Conclusions}
Tropical geometry tools were used in this paper to present a methodology to analyze the C-space of a mechanism by determining Puiseux series of the 1-dimensional algebraic varieites at a given point. The real and complex transversal branchings in the configuration curve of two foldable four bar mechanisms were determined. The cusp in the C-space of a double Watt mechanism was detected along with two other branches, where the mechanism had smooth motions.\\
Since the C-space was described in terms of Study parameters, the double Watt mechanism at $S$ had a branching of type 6 in Table~1 of~\cite{Custodio2020}. As the cuspidal intersections are succesfully detected by this approach, attempts to detect other types of branchings~\cite{Custodio2020} will be made in the future. 

\bibliographystyle{spmpsci}
\bibliography{bibfile}
\end{document}